\documentclass{article}

\usepackage{arxiv}

\usepackage[round]{natbib}
\usepackage[utf8]{inputenc} % allow utf-8 input
\usepackage[T1]{fontenc}    % use 8-bit T1 fonts
\usepackage{hyperref}       % hyperlinks
\usepackage{url}            % simple URL typesetting
\usepackage{booktabs}       % professional-quality tables
\usepackage{amsfonts}       % blackboard math symbols
\usepackage{nicefrac}       % compact symbols for 1/2, etc.
\usepackage{microtype}      % microtypography
\usepackage{times}
\usepackage{latexsym}
\usepackage{amssymb}
\usepackage{amsthm}
\usepackage{amsmath}
\usepackage{graphicx}
\usepackage{multirow}
\usepackage[dvipsnames]{xcolor}
\usepackage[titlenumbered,ruled]{algorithm2e}
\usepackage{float}
\usepackage{graphicx}
\graphicspath{ {figs/} }
\usepackage{wrapfig}
\usepackage{url}

\newcommand{\nil}[1]{}
\newcommand{\fw}[1]{}
\newcommand{\jb}[1]{}
\newcommand{\kqw}[1]{}
\newcommand{\lw}[1]{}
\newcommand{\boyi}[1]{}

\newcommand{\comment}[1]{}

\renewcommand{\vec}[1]{\boldsymbol{\mathbf{#1}}}

\def\ba{\vec{a}}

\def\bc{\vec{c}}

\def\bq{\vec{q}}

\def\pt{P_\theta}

\newcommand{\method}{\textit{Integrated Triaging}}

\newcommand{\methodbert}{Triaged-BERT}
\newcommand{\methodffn}{Triaged-FFN}
\newcommand{\moduleA}{Triage Module}
\newcommand{\moduleB}{Expert Model}

\newcommand{\model}{\method{}}

\definecolor{ffn}{HTML}{2872ea}
\definecolor{triagedffn}{HTML}{48a3ea}
\definecolor{bert}{HTML}{ff9558}
\definecolor{triagedbert}{HTML}{f56666}
\definecolor{candidate}{HTML}{5E91FF}
\definecolor{answer}{HTML}{293FAB}
\definecolor{sentence}{HTML}{FF4948}

\title{Integrated Triaging for Fast Reading Comprehension}

\author{%
  Felix Wu, Boyi Li, Lequn Wang \\
  Cornell University \\
  \And
  Ni Lao \\
  SayMosaic Inc.\\
  \And
  John Blitzer \\
  Google Inc.
  \And
  Kilian Q. Weinberger \\
  Cornell University
}

\begin{document}

\maketitle

\begin{abstract}
Although according to several benchmarks automatic machine reading comprehension (MRC) systems have recently reached super-human performance, less attention has been paid to their computational efficiency. 
However, efficiency is of crucial importance for training and deployment in real world applications.
This paper introduces \textit{\method{}}, a framework that prunes almost all context in early layers of a network, leaving the remaining (deep) layers to scan only a tiny fraction of the full corpus. 
This pruning drastically increases the efficiency of MRC models and further prevents the later layers from overfitting to prevalent short paragraphs in the training set.
Our framework is extremely flexible and naturally applicable to a wide variety of models.
Our experiment on doc-SQuAD and TriviaQA tasks demonstrates its effectiveness in consistently improving both speed and quality of several diverse MRC models.

\end{abstract}

\section{Introduction}

Machine reading comprehension (MRC) has seen impressive advances in recent years. These have been fueled by innovations in deep learning~\citep{Hochreiter1997LongSM,cho2014gru,bahdanau2014neural}, but progress is also in no small part due to the creation of standardized benchmark data sets, e.g. SQuAD~\citep{rajpurkar2016squad}.
Since its introduction, results on the SQuAD task have continuously improved, with at least one model, BERT~\citep{devlin2018bert}, achieving superhuman performance.  

Although these levels of accuracy are certainly impressive, the top performing models on the SQuAD leaderboard tend to be too slow for practical purposes. An ideal MRC system needs to scan an enormous number of trusted documents at rapid speed to find and return a correct answer within a timely manner. 
At run-time in real-world applications, where humans may be waiting patiently for the answer, this is of uttermost importance. 

However, speed is also a crucial factor during training. 
A more realistic variant of SQuAD is doc-SQuAD \citep{Clark2017SimpleAE}, which provides the MRC system with the whole document, sometimes containing hundreds of paragraphs. Most models are too slow to be trained on such a large input context, and therefore researchers train their models on the SQuAD data set instead, which is reduced to a few ``golden paragraphs'' that are known to contain a correct answer. Alas,  models that are trained on SQuAD golden paragraphs typically do not fare well in the doc-SQuAD setting. 
The unexpected additional text during testing entails a  covariate shift (the change of the data distribution from the training set to the test set) and yields incorrect answers as the models do not generalize to longer contexts.

\begin{wrapfigure}{R}{0.5\textwidth}
    \centering
    \includegraphics[width=0.48\textwidth]{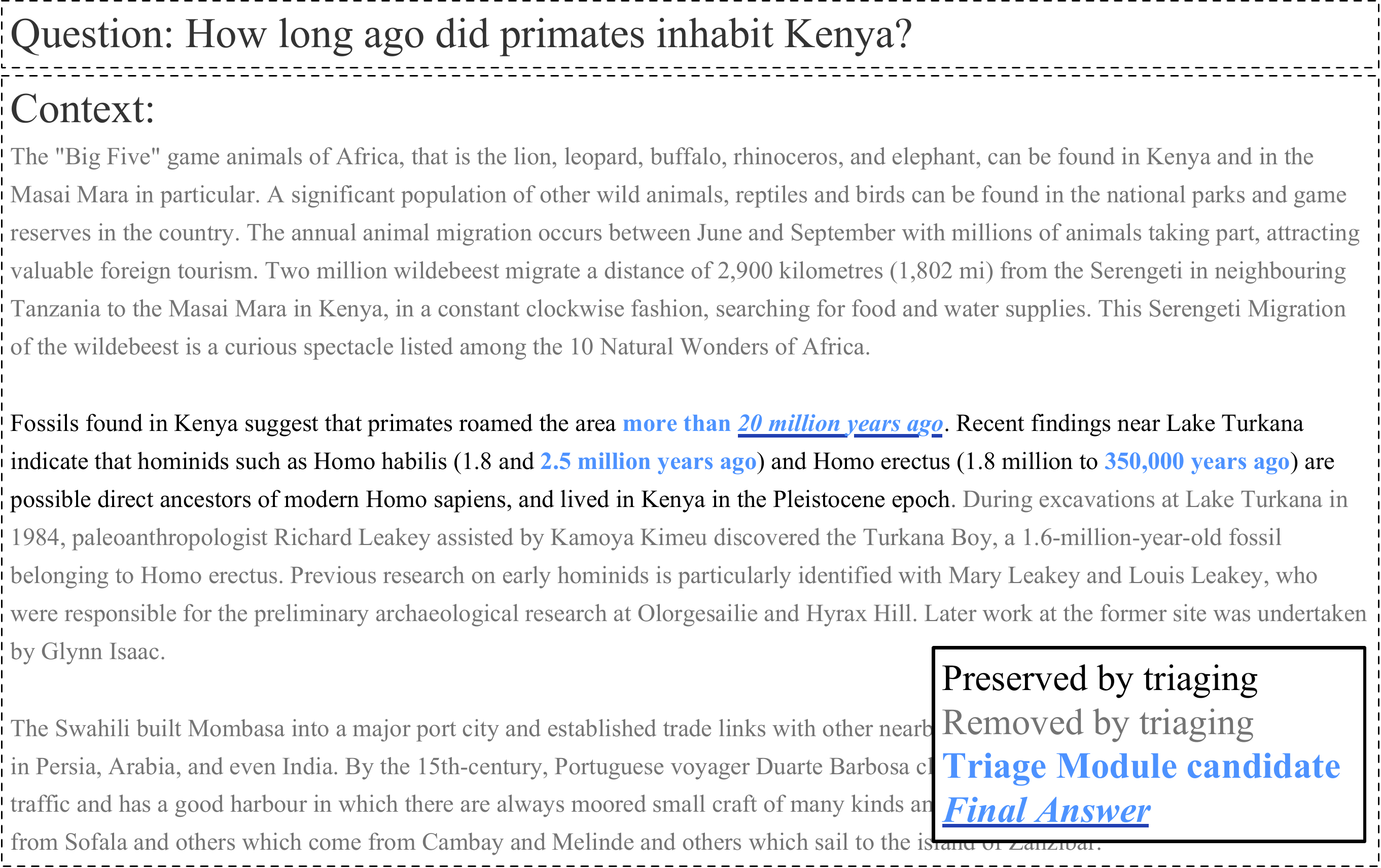}
    \caption{The \moduleA{} generates {\color{candidate}\textbf{answer candidates}} and keeps only \textbf{the features of the sentences they belong to} for the later part of the model to produce the {\color{candidate}\textbf{\textit{\underline{final answer}}}}.}
    \label{fig:example}
\end{wrapfigure}

In this paper we propose \method{}, a small but significant   modification applicable to most existing MRC models. We propose to insert a light-weight \emph{\moduleA{}}  between two early layers of an MRC net. This \moduleA{} replicates the final answer layer of the MRC architecture and generates a probability score over plausible answer sub-sequences for any context passage it is applied to. Most importantly, the confidence of this probability score determines whether the latent features of this passage should be passed on to the full model as a plausible answer candidate, be removed as irrelevant, or---in cases of very high confidence---immediately returned as the final solution. 
See \autoref{fig:example} for an illustrative example. 
Because the \moduleA{} successfully removes most irrelevant passages from the input context, training on the full doc-SQuAD becomes similar to training on the golden paragraph of SQuAD. 
Consequently, with a \moduleA{}, models do not suffer from  covariate shift and  
can be trained efficiently on golden paragraphs alone, but during testing scale naturally to \textit{full document} context lengths. 

During inference when the \moduleA{} is confident enough with its answer, the model stops the inference procedure and outputs this answer, which is the so called \textit{early-exit} mechanism~\citep{shen2017reasonet,huang2017multi,wangresource,das2018multistep}.
Otherwise, the \moduleA{} keeps the features of the relevant context, removes the rest, and passes them to the following layers of the model, which we refer to as the \textit{context-pruning} mechanism.
Our experiments show that these two mechanisms speed up the inference time of the model.

We evaluate \textit{\method{}} on three primary benchmarks.  First, we show that it can achieve state-of-the-art results on doc-SQuAD when applied to two reading comprehension architectures. 
When applied to BERT~\citep{devlin2018bert}, it improves the F1 score by 1.8 with a $20\%$ reduction of inference time.
When applied to FastFusionNet~\citep{wu2019fastfusionnet}, an efficient but less accurate model, it gives a $4.3$ improvement in F1 score with a $20\%$ speedup.
Second, on TriviaQA~\citep{joshi2017trivia} where the golden paragraphs are not provided, \textit{\method{}} results in a 80\% speedup with a little drop in performance. 
Third, \method{} tops the DAWNBench \citep{coleman2017dawnbench} leaderboard in inference time.
Despite its rapid speed, it removes over up to $95\%$ of the context on document level test cases while reliably keeping the relevant answer sections.
Compared to the previous state-of-the-art on doc-SQuAD, \method{} is substantially faster to train, achieves a significantly higher F1-retrieval score. 
Including the gain from using a faster MRC model, our fastest model enjoys a $10\times$ speedup compare to the previous state-of-the-art model while maintaining the same accuracy.

\section{Background and Related Work}
\subsection{Machine Reading Comprehension}

The task of machine reading comprehension (MRC) is to extract a section of a given context as the answer to an input question that is provided in plain text. 
Formally, the goal is to pick a contiguous span $[b,e] (1 \le b \le e \le n)$ from
a context text $\bc = \{c_1,  \dots, c_n\}$ with $n$ tokens as the answer to a question $\bq = \{q_1,  \dots, q_m\}$ with $m$ tokens~\citep{rajpurkar2016squad}.
A popular approach is to use a pointer network \citep{vinyals2015pointer} style output module to 
assign a distribution over the context words representing the probability that a word is the beginning of the answer span $\pt(b|\bq, \bc)$, and similarly to get the end prediction $\pt(e|b, \bq, \bc)$. 
The model uses exhaustive search within a limited range 
to find the highest scoring span $[b^*, e^*] = \mathrm{argmax}_{1 \leq b \leq e \leq n}\pt(b, e|\bq, \bc)$. Sometimes, an additional constraint $e-b < l$ is added to restrict the length of the answer span not surpassing $l$ tokens. 

{The Stanford Question Answering Dataset (SQuAD)}~\citep{rajpurkar2016squad} is one of the most popular reading comprehension datasets and contains over 100K question-answer-passage tuples (87K for training, 10K for development, and 10K for test). The dataset was labeled by crowdsource workers who, given a paragraph, were asked to generate questions based on it. For each passage another group of workers attempted to highlight a span in the passage as the answer. This ensures that the passage also contains sufficient information and the answer is always present. 
There are three flavors of this task:
1) {\em SQuAD}~\citep{rajpurkar2016squad} the model is presented with a ``golden'' paragraph, which contains the answer;
2)  {\em document SQuAD}~\citep{Clark2017SimpleAE}  the model is presented with the full relevant document consisting of many paragraphs, one of which contains the answer;
3)  {\em corpus SQuAD}~\citep{chen2017reading} the model is presented with the whole corpus (including all documents).
Our main result is on document SQuAD where the efficiency of machine reading is more prominent than SQuAD. 
However, the architecture we introduce readily also applies to corpus SQuAD.

The state-of-the-art approach on doc-SQuAD is to train a model with the golden paragraph and several that are randomly selected (drawn from top-$k$ paragraphs similar to the question based on TF-IDF cosine similarity)~\citep{Clark2017SimpleAE}. During inference, the model uses  TF-IDF cosine similarity to select the top-$k$ most relevant paragraphs, applies them to the model and normalizes their predictions, often referred to as the shared-norm method.
Here, the randomly selected paragraphs slow down the training procedure and TF-IDF, which is not trained, has to be used with precaution (i.e. a large $k$ value) to avoid discarding correct answers---slowing down test inference on longer documents.    

\subsection{MRC Models}
Despite the recent impressive advances in performance, the high-level MRC model layouts have remained mostly the same~\citep{wang2017matchlstm,seo2016bidirectional,xiong2017dcn+,Weissenborn2017fastqa,chen2017reading,hu2018reinforced,huang2018fusionnet,wei2018qanet,wu2019fastfusionnet,wang2018multi,devlin2018bert}: An MRC model comprises an input layer, a series of encoding layers, and an output layer.
An MRC model takes as input a question $\bq$ and a context $\bc$ and estimates the probability of an answer span  $\pt(b, e|\bq, \bc)$.

\paragraph{Input layer.}
The input layer contains a word embedding module, usually initialized with pre-trained word vectors~\citep{pennington2014glove}. Some~\citep{seo2016bidirectional,peters2018elmo} use character embeddings followed by a character-level ConvNet to generalize to unseen words, some~\citep{McCann2017CoVe,huang2018fusionnet,peters2018elmo,wang2018multi} use pre-trained contextual embedding features, and others~\citep{chen2017reading,huang2018fusionnet,hu2018reinforced} use addition word features such as part-of-speech tags, named entity recognition tags, term frequency, or whether a context word is in the question. The outputs of this layer are two sequences of vectors $\mathbf{Q}^0 = \{\bq_1^0, \dots,\bq_m^0\}$ and $\mathbf{C}^0 = \{\bc_1^0, \dots, \bq_n^0\}$ of the question and the context, respectively.

\paragraph{Encoding layers.}
The $i$-th encoding layer takes as inputs the representations of the question and the context from the previous layer $\mathbf{Q}^{i-1}$ and $\mathbf{C}^{i-1}$, refines them, and produces $\mathbf{Q}^i$ and $\mathbf{C}^i$. This is usually done with recurrent layers~\citep{Weissenborn2017fastqa,chen2017reading}, convolutional layers~\citep{wu2017fast,wei2018qanet}, attention modules~\citep{seo2016bidirectional,Chen2017SmarnetTM,xiong2017dcn+,wang2017rnet,huang2018fusionnet,wang2018multi,devlin2018bert}, or point-wise feed-foreward networks~\citep{liu2018san}.
The question and the context can be encoded in parallel, or interactive with each other. Since the context is longer and requires deeper understanding, some models have more layers of operations for the context than the question.
Here, we treat these cases as identity mappings, i.e. $\mathbf{Q}^i = \mathbf{Q}^{i-1}$ and assume each layer incorporates both, the query and the context embedding. 

\paragraph{Output layers.}
The output layer produces the answer span $[b, e]$ using $\mathbf{Q}^{L}$ and $\mathbf{C}^{L}$ from the last encoding layer. Inspired by PointerNet~\citep{vinyals2015pointer}, some~\citep{chen2017reading,huang2018fusionnet,liu2018san} summarize $\mathbf{Q}^{L}$ into a vector and employ a bilinear attention to derive the answer; others~\citep{seo2016bidirectional,devlin2018bert} ignore $\mathbf{Q}^{L}$ and rely on only $\mathbf{C}^{L}$ to find the answer assuming that the question information have been encoded into $\mathbf{C}^{L}$ through the previous layers.
In pursuit of high performance, \citep{seo2016bidirectional, wei2018qanet} use heavy output layers that refines $\mathbf{C}^{L}$ with multiple neural layers conditioned on the start prediction before outputting the end position. However, the success of \citep{devlin2018bert} shows that predicting the start and end independently still works.

\subsection{Efficient Inference}
\paragraph{Coarse-to-fine inference.}
\citep{lee2015deeply} proposed deep supervision which saves computation for ``easy'' examples through early-exits while still obtaining good predictions for ``difficult'' examples in the later layers~\citep{shen2017reasonet,huang2017multi,wangresource,das2018multistep}.  
Arguably, most similar to our work may be \citep{choi2017coarse} and \citep{gehrmann2018bottom}. \citep{choi2017coarse} proposes a two-stage method that relies on a sentence selection model to filter the context and execute a question answering model on the selected sentence.
Their method requires a second network trained either separately or jointly for a sentence selection task. Additionally, their MRC model processes the selected context from scratch again instead of re-using the extracted features.
Re-using features benefits the model in two ways: i) saving computation ii) preserving context information from the surrounding pruned sentences.  Both their and our methods require knowing the gold sentence during training which can also be approximated by distant supervision, i.e., taking the first sentence with the ground truth answer as the gold sentence.
 However, their training procedure is based on the whole document, making it substantially slower than ours.  Additionally, their methods use a fixed size sentence vector, while our \moduleA{} produces a sequence of word representations, which allows the \moduleB{} to make more accurate predictions.
For abstractive machine summarization, \citep{gehrmann2018bottom} proposes to generate a mask that prevents the PointerNet-like~\citep{vinyals2015pointer} decoder from copying irrelevant tokens from an article.

\paragraph{Phrase-indexed Question Answering} (PIQA)
\citep{seo2018phrase} is another framework of efficient reading comprehension, where the feature extraction of contexts is isolated from the encoding of questions, i.e. $\mathbf{C}^i$ doesn't depend on $\mathbf{Q}^{i-1}$ in all encoding layers, in order to reuse the context features and trade storage for efficiency.
However, PIQA introduces a significant performance drop ($30\%$ F1 score) compared to conventional QA models, which suggests that question information plays a vital role in reasoning contexts.
Moreover with our extremely efficient \moduleA{}, an experiment shows that re-computing the features is about two times faster as loading the pre-computed features from local disk to memory; suggesting that the pre-computation may not be necessary.

\section{Integrated Triaging}
\begin{figure*}[t]
    \centering
    \includegraphics[width=0.8\textwidth]{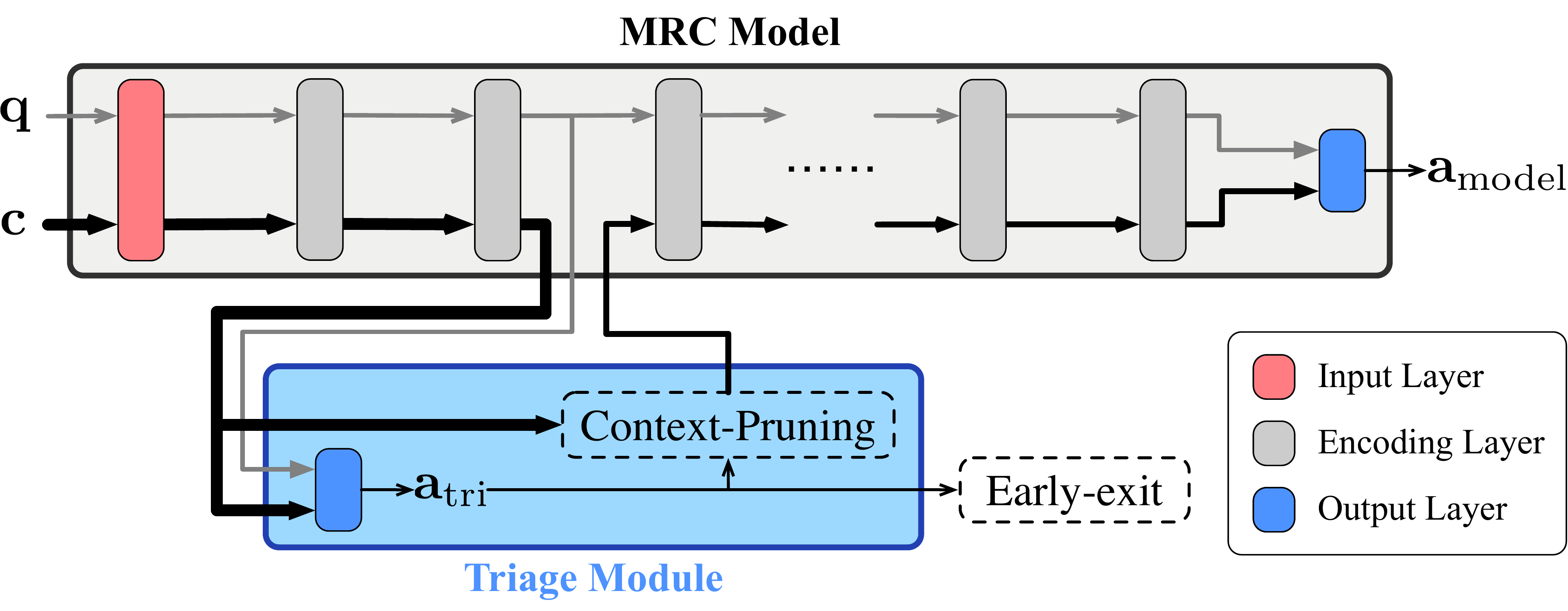}
    \caption{
    An illustration of the changes implied through \method{}. A Triage Module is inserted inbetween the early layers a standard MRC network (consisting of input, encoding, and output layers). The Triage Module consists of a replication of the output layer, followed by context pruning and early-exit. $\mathbf{q}$: the question, $\mathbf{c}$: the context, $\mathbf{a}$: the answer (model / triaging).
    }
    \label{fig:arch}
    \vspace{-0.2in}
\end{figure*}

\method{} adds an efficient \moduleA{}
to an accurate (but slow) MRC model to speed up inference through two mechanisms: \textit{early-exit} and \textit{context-pruning}. (See \autoref{fig:arch} for a schematic illustration.)

\paragraph{\moduleA{}.}
Given an MRC model, we replicate its output layer (without weight sharing) as the core component of the \moduleA{}. As illustrated in  \autoref{fig:arch}, we attache this replicated output layer onto the output of the $T^{th}$ earliest layer. The answer $\ba_{\mathrm{tri}}$ consists of a probability distribution over answer sections ($b$ and $e$). 
During inference, the model sees long documents with several paragraphs in them as the context.
Most of the context neither contains the answer nor provides supporting information for deducing the answer. Removing irrelevant context not only reduces computation but also guard the answering layer from distractions.
See \autoref{alg:inference} for a summarization of the inference procedure in pseudo-code.
We first execute the input layer, the first $T$ encoding layers, and the \moduleA{}, and return $\ba_\mathrm{tri}$ if the module is confident enough (the score of the most plausible answer is greater than a preset threshold $t$). Otherwise, the context-pruning is applied to keep context features of the top $K$ answer candidates as well as the sentences they belong to. 
These pruned context features are then passed on to the subsequent (deeper) layers of the model with the question features to generate the final answer. Based on the confidence of these output probabilities, a passage is either passed on, pruned, or output as the final answer.

\begin{wrapfigure}{R}{0.5\textwidth}
    \begin{minipage}{0.5\textwidth}
        \begin{algorithm}[H]
        \small
        \caption{\method{}}
        
        \SetKwInOut{Input}{Input}
        \SetKwInOut{Output}{Output}
        \Input{
        context $\bc$, question $\bq$, early-exit threshold $t$, context-pruning number of candidates $K$.
        }
        \Output{predicted answer span $\mathbf{s} = [b, e]$}
        \textbf{Notations:}  shared input layer $\mathrm{IN}$, 
        $i$-th encoding layer $\mathrm{ENC}_i$, 
        output layer of the \moduleA{} $\mathrm{OUT_{tri}}$, 
        output layer of the MRC model $\mathrm{OUT_{\mathrm{model}}}$\\
        $(\mathbf{Q}^0, \mathbf{C}^0) \leftarrow \mathrm{IN}(\bq, \bc)$\;
        \For{$i\gets1$ \KwTo $T$}{
            $(\mathbf{Q}^{i}, \mathbf{C}^{i}) \leftarrow \mathrm{ENC}_i(\mathbf{Q}^{i-1}, \mathbf{C}^{i-1})$\;
        }
        $\mathbf{a}_{\mathrm{tri}} \leftarrow \mathrm{OUT_{tri}}(\mathbf{Q}^T, \mathbf{C}^T)$\;
        \eIf{$\max{\mathbf{a}_{\mathrm{tri}}} > t$}{
            $\mathbf{s} \leftarrow \mathrm{argmax}(\mathbf{a}_{\mathrm{tri}})$ 
            \tcp*{Early-Exit} 
        }{
            $\tilde{\mathbf{C}}^{T} \leftarrow \mathrm{context\_pruning}(\mathbf{C}^T, \mathbf{a}_{\mathrm{tri}})$ \;
            \For{$i \gets T+1$ \KwTo $L$}{
                $(\mathbf{Q}^{i}, \tilde{\mathbf{C}}^{i}) \leftarrow \mathrm{ENC}_i(\mathbf{Q}^{i-1}, \tilde{\mathbf{C}}^{i-1})$\;
            }
            $\mathbf{a}_{\mathrm{model}} \leftarrow \mathrm{OUT_{\mathrm{model}}}(\mathbf{Q}^L, \tilde{\mathbf{C}}^{L})$\;
            $\mathbf{s} \leftarrow \mathrm{argmax}(\mathbf{a}_{\mathrm{model}})$ 
        }
        
        \label{alg:inference}
        \end{algorithm}
    \end{minipage}
\end{wrapfigure}

\paragraph{Context-Pruning.}
The Context-Pruning module keeps
all sentences containing the top $K$ answer span candidates with the highest confidence. 
Given the triage answer $\mathbf{a}_{\mathrm{tri}} = P_{\theta_{\mathrm{tri}}}(b_{\mathrm{tri}}, e_{\mathrm{tri}}|\mathbf{q}, \mathbf{c})$, we find the top $K$ answer candidates $[b_1, e_1], \dots, [b_K, e_K]$ with the highest scores and the sentences they belong to. We keep the features of the tokens in these sentences and prune everything else. The order of kept features are preserved. It is possible that all top $K$ answer candidates are in the same sentence and we keep only this sentence (letting the later layers decide the exact beginning and end of the answer).
Formally, the context-pruning module takes  $\mathbf{C}^T$, the context features after the $T$-th encoding layer, and  the triage answer $\mathbf{a}_{\mathrm{tri}}$ as inputs, 
and outputs the pruned context  $\mathbf{\tilde{C}}^{T} = \{\mathbf{c}^T_{p_1}, \dots,\mathbf{c}^T_{p_k}\}$ with $k$ tokens at positions $1 \leq p_1 < \dots < p_k \le n$.

\paragraph{Training.}
During training the model is presented with questions and paragraph-level contexts, and produces two answers: $\ba_\mathrm{tri}$ (the answer from the \moduleA{}) and $\ba_\mathrm{model}$ (the answer from the MRC model).
Early-exit and context-pruning are disabled and the two output layers see the same short context. 
We simply minimize the sum of the negative log-likelihood of both answers with respect to the gold answer $\ba^*$:
\[
\mathcal{L}_{\mathrm{all}} = 
\mathcal{L}_{\mathrm{nll}}(\mathbf{a}^*, \mathbf{a}_{\mathrm{tri}}) + \mathcal{L}_{\mathrm{nll}}(\mathbf{a}^*, \mathbf{a}_{\mathrm{model}})  
\]

\section{Triaged MRC Models }
%Applications}
\label{sec:impl}
In this section we briefly discuss how we apply \method{} to two sample MRC architectures, BERT~\citep{devlin2018bert} and FastFusionNet (FFN)~\citep{wu2019fastfusionnet}. The former is the most accurate and the later the most efficient model on SQuAD 1.1.
We set  $T$  (the number of layers before the \moduleA{}) to the smallest value such that the \moduleA{} can achieve an F1 score of $75\%$ on the SQuAD dev-set, which is considered as an acceptable performance in DAWNBench~\citep{coleman2017dawnbench}.

\subsection{\methodbert{}}
\begin{figure}[t]
    \centering
    \begin{minipage}{0.45\textwidth}
        \centering
        \includegraphics[width=\textwidth]{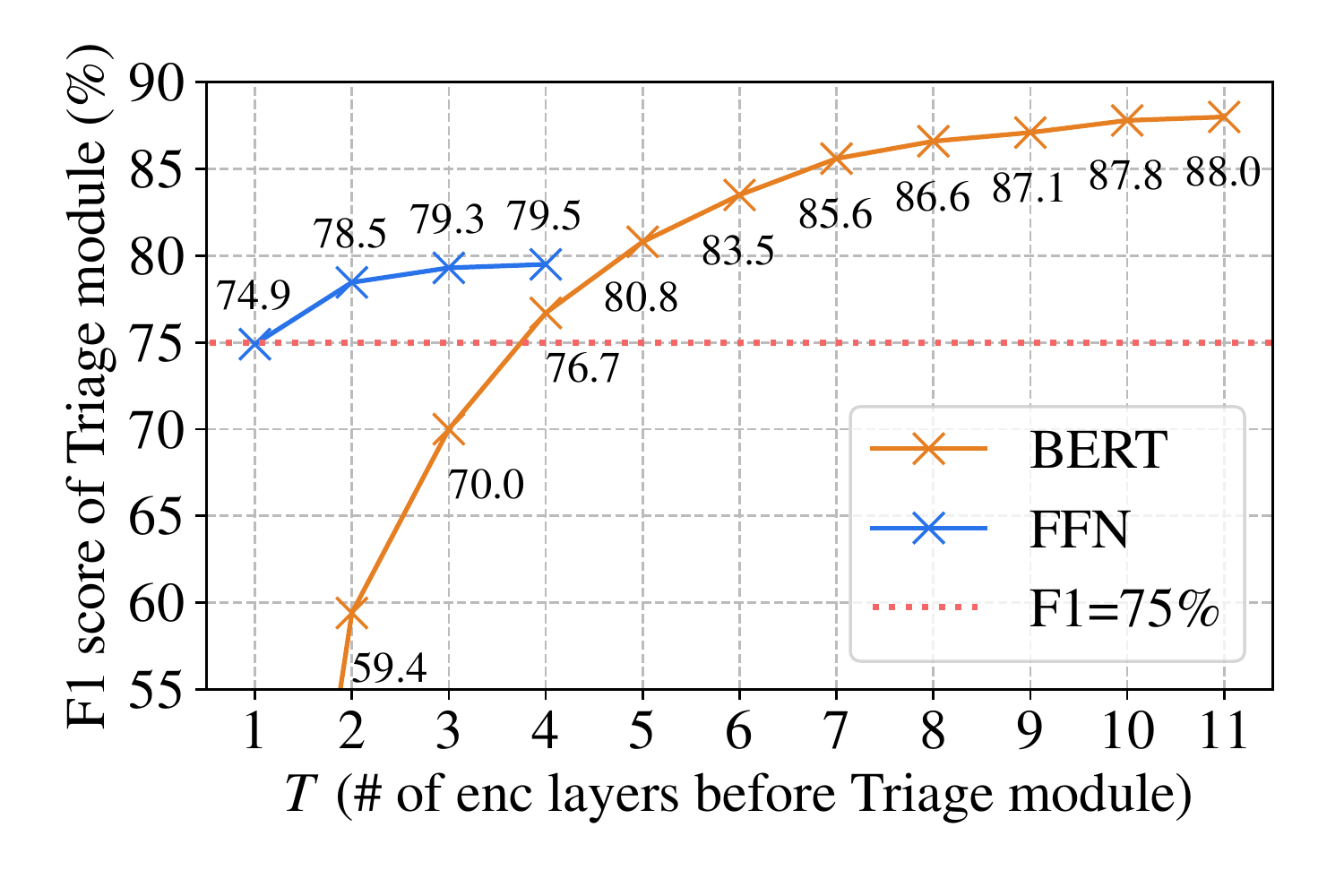}
        \vspace{-0.1in}
        \caption{We choose the value of $T$ such that the \moduleA{} achieves an F1 score of at least $75\%$ on the SQuAD dev set, 
        resulting in $T\!=\!4$ for BERT and $T\!=\!2$ for FFN. We do not evaluate FFN beyond $T\! =\! 4$, as $T\!=\!2$ is already sufficient.
        }
        \label{fig:layers}
    \end{minipage}\hfill
    \begin{minipage}{0.45\textwidth}
        \centering
        \includegraphics[width=\textwidth]{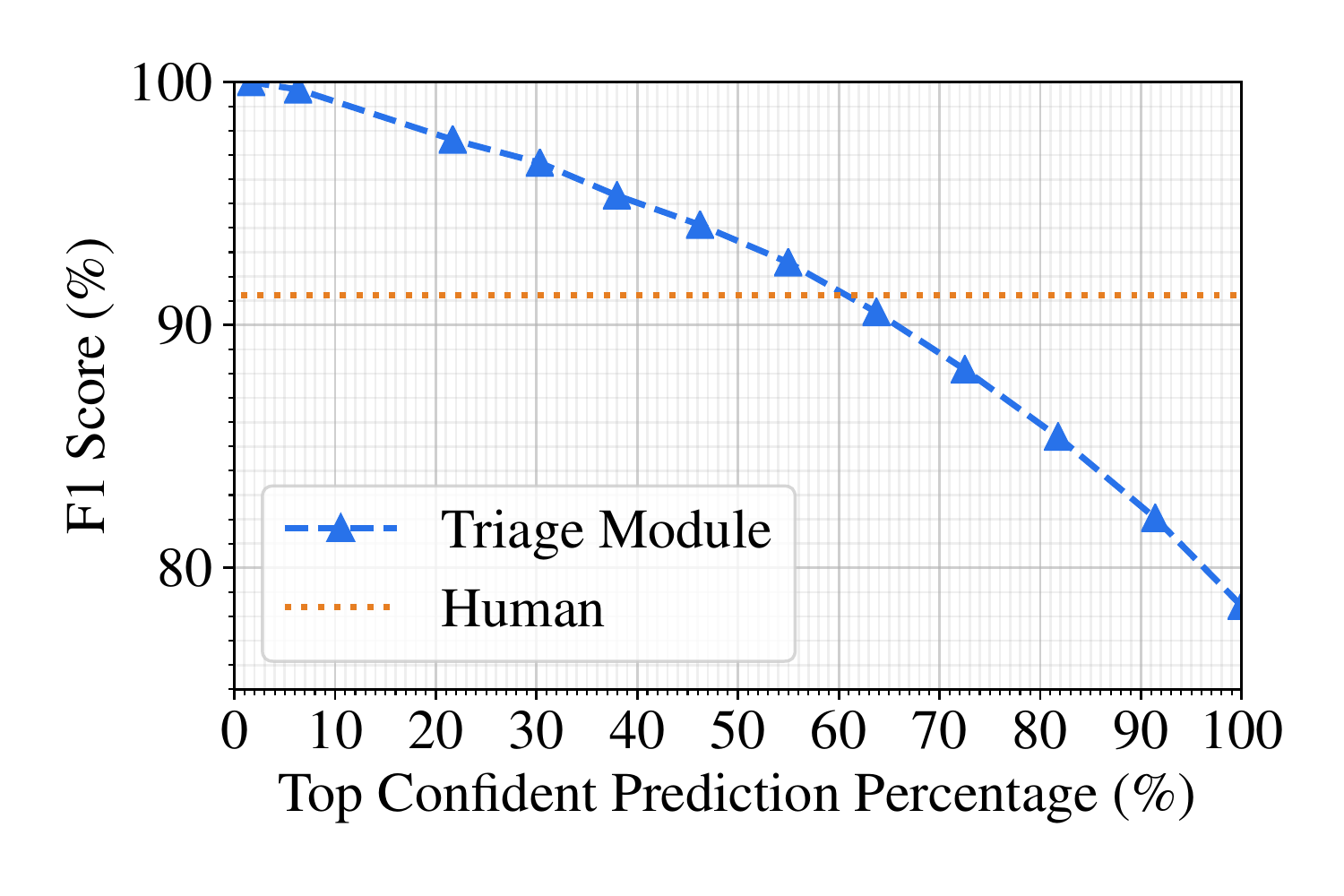}
        \vspace{-0.1in}
        \caption{The F1 score corresponding to the top $N \%$ most confident predictions of the FFN-\moduleA{} on the SQuAD dev set. 
        The \moduleA{} matches human performance for the top $60 \%$ most confident predictions. It is therefore safe to use the full model for less than $40\%$ of the data. 
        }
        \label{fig:triage_f1}
    \end{minipage}
    \vspace{-0.2in}
\end{figure}

BERT (Bidirectional Encoder Representations from Transformers~\citep{devlin2018bert}) is the state-of-the-art model on the SQuAD public leaderboard. The backbone of this model is pre-trained on the masked language modeling and a next sentence prediction tasks. 
BERT concatenates the question and context wordpiece tokens, i.e. byte pair encoding \citep{gage1994new}, as the inputs and uses the pre-trained Transformer blocks~\citep{vaswani2017attention} (composed of self-attention and point-wise feed-forward layers) to refine the representations of the question and the context.
The answer prediction module is a simple point-wise linear layer, which projects the outputs of the last layer to two dimensions (one for the start and one for the end), softmax-normalizes the logits across the context sequence positions, and predicts the start and end independently.
We treat each Transformer block as an encoding and use BERT-base which has 12 Transformer blocks.
Figure~\ref{fig:layers} shows the performance of the \moduleA{} under varying values for $T$.
As we can observe, 4 pre-trained transformer blocks (out of 12) are sufficient to reach $75\%$ F1 score; thus, we set $T=4$ throughout the rest of the experiments.

\subsection{\methodffn{}}
FastFusionNet (FFN)~\citep{wu2019fastfusionnet} is an efficient machine reading comprehension model.
FFN requires 8 layers of bidirectional simple recurrent unit (BiSRU)~\citep{lei2017sru} as well as two Fully-aware attention layers  for contexts, and 6 layers of BiSRU for questions.
Its output layer uses attention mechanism to summarize the question sequence into a vector and employs a bi-linear attention to generate the prediction of the start position. Conditioned on this start prediction, it updates the question summary vector and applies another bi-linear attention to predict the end position.
Each of the first four encoding blocks contains two 1-layer BiSRUs: one for the context and the other for the question.
\autoref{fig:layers} shows that $T=2$, i.e. having two layers of SRUs for both questions and contexts, is sufficient.
There are many choices of how we view the later layers.
However, we don't evaluate settings with $T\!>\!4$ since we have met the requirement of F1 $\ge 75\%$.
Please refer to the original paper~\citep{wu2019fastfusionnet} for additional details on the model.

To evaluate the efficacy of the \moduleA{} in the context of early-exiting, we analyze the performance of the \moduleA{} on the examples it is most confident about.
\autoref{fig:triage_f1} shows that this \moduleA{} can match human performance on the top $60 \%$ confident examples, and only the remaining 40\% are sufficiently hard to require a ``second opinion'' from the full FFN.
We show in section~\ref{sec:exp} that this approach can reduce the average inference time significantly with no negative impact on  F1 score.

\section{Experiments}
\label{sec:exp}

In this section, we 
demonstrate that triaged MRC models outperform
the state-of-the-art solution~\citep{Clark2017SimpleAE} and other baselines on the doc-SQuAD task. 
To show the trade-offs among different parameter choices, we conduct an ablation study where we control the number of paragraphs read by the models. 
We also show that our method works with distant supervision on TriviaQA without golden paragraphs provided.
Finally, we illustrate the efficiency of our fast models based on the DAWNBench~\citep{coleman2017dawnbench} evaluation. The experiments of TriviaQA and DAWNBench are provided in the supplemental materials.

\subsection{Experimental Setup}
We briefly describe the experimental setup in this section. More details are revealed in the supplemental materials.
\paragraph{Datasets}
The original SQuAD v1.1 dataset is composed of a training set, a development set, and a holdout test set. Since the official test set is held secret, we split the development set into a validation set (\textit{dev-val}) with the first 16 documents and 4306 questions and a test set (\textit{dev-test}) with the remaining 32 documents and 6264 questions. We use \textit{dev-val} for hyper-parameter tuning and the ablation study, while \textit{dev-test} is held for the comparison with the baseline models. We also evaluate our method on TriviaQA \citep{joshi2017trivia} which is also a span prediction reading comprehension dataset but without golden paragraphs.

\paragraph{Baselines}
One common strategy on full document MRC is to use TF-IDF cosine similarity between the question text and a paragraph to pre-screen potential answer candidates. Although not very accurate, TF-IDF is very fast and tends to be effective at removing completely irrelevant paragraphs.  
For instance, \citep{Clark2017SimpleAE} evaluates their model with up to the top 15 paragraphs. We use their TF-IDF paragraph ranker for all baselines and our models (as initial pre-screening). 

We consider the following baselines: i) the pipeline sentence selection method introduced in \citep{choi2017coarse} with various sentence selector models, ii) BiDAF + Self-attn + MPT (multi-paragraph training method proposed in \citep{Clark2017SimpleAE}), iii) FFN, and iv) BERT.
Details of the baselines are provided in the supplemental materials.

\begin{table*}
\centering \small
\resizebox{\linewidth}{!}{%
\begin{tabular}{l|c|rr|rr|rrr|r}
\multirow{2}{*}{Model} & \multirow{1}{*}{\# of}  & \multicolumn{2}{l|}{\textit{dev-val}} & \multicolumn{2}{l|}{\textit{dev-test}} & \multicolumn{3}{l}{\textit{dev-test} Latency (ms)} 
&\multicolumn{1}{|l}{Pruned}
\\
 &  \multicolumn{1}{l|}{para} & \multicolumn{1}{l}{EM} & \multicolumn{1}{l|}{F1} & \multicolumn{1}{c}{EM} & \multicolumn{1}{c|}{F1} & \multicolumn{1}{c}{Avg.} & \multicolumn{1}{c}{90\%} & \multicolumn{1}{c}{99\%}&\multicolumn{1}{|l}{Portion} \\ \hline
Pipeline (BoW) & 3 & 53.5 & 60.3 & 58.2 & 65.5 & 8.2$\pm$0.12 & 9.3$\pm$0.09 & 15.6$\pm$0.80  & 28.5\% \\
Pipeline (CNN) & 3 & 53.1 & 59.9 & 56.2 & 63.5 & 7.7$\pm$0.43 & \textbf{8.6$\pm$0.61} & \textbf{11.0$\pm$2.59} & 29.4\% \\
Pipeline (SRU) & 3 & 53.3 & 60.0 & 58.1 & 65.6 & 9.4$\pm$0.62 & 10.3$\pm$0.59 & 12.0$\pm$0.74 & 29.8\%\\ \hline
Clark \& Gardner (2017)~\citep{Clark2017SimpleAE} & 3 & 60.5 & 68.1 & 64.7 & 73.6 & 61.6 & 70.3 & 85.9 & 0\% \\ 
Clark \& Gardner (2017)~\citep{Clark2017SimpleAE} & 8 & 62.1 & 69.8 & 65.7 & 74.5 & 63.4$\pm$0.48 & 73.9$\pm$0.48 & 92.2$\pm$1.04 & 0\% \\ 
\hline
FFN & 3 & 57.2 & 64.3 &  62.3 & 70.3 & 8.3$\pm$0.32 & 9.4$\pm$0.42  & 14.2$\pm$2.47 & 0\%\\
\color{blue} \methodffn{} ($K = 20, t = 0.4$)  &\color{blue} 3 &\color{blue} 61.5  &\color{blue} 69.5 &\color{blue}  65.8 &\color{blue} 74.6 &\color{blue} \textbf{6.2$\pm$0.07}  &\color{blue} 9.4$\pm$0.17 &\color{blue} 11.6$\pm$0.26 & 90.5\%\\
\color{blue} \methodffn{} ($K = 50, t = \infty$) &\color{blue} 20 &\color{blue} 62.8  &\color{blue} 70.8 &\color{blue}  67.5 &\color{blue} 76.2 &\color{blue} 24.1$\pm$0.04  & \color{blue} 34.1$\pm$0.02 &\color{blue} 95.4$\pm$0.25 & 93.4\% \\
BERT  & 3 & 66.3 & 72.3 & 69.3 & 76.2 & 30.1$\pm$0.11 & 40.1$\pm$0.09 & 60.2$\pm$0.25 & 0\%\\
\color{blue}\methodbert{} ($K = 20, t = \infty$) &\color{blue} 3 &\color{blue} 67.7 &\color{blue} 73.6 &\color{blue} \textbf{71.0} &\color{blue} \textbf{78.0} &\color{blue} 23.9$\pm$0.26 &\color{blue} 28.2$\pm$0.40 &\color{blue} 40.9$\pm$1.08 & 79.5\% \\
\end{tabular}
}
\caption{Main results on doc-SQuAD. 
$K$: \# of candidates to keep, $t$: early-exit threshold, $t = \infty$: no early-exit.
For latency, we report the mean and standard deviation over 5 runs.
}
\label{tbl:main_results}
\end{table*}

\begin{wrapfigure}{R}{0.48\textwidth}
    \centering
    \vspace{-2ex}
    \includegraphics[width=0.45\textwidth]{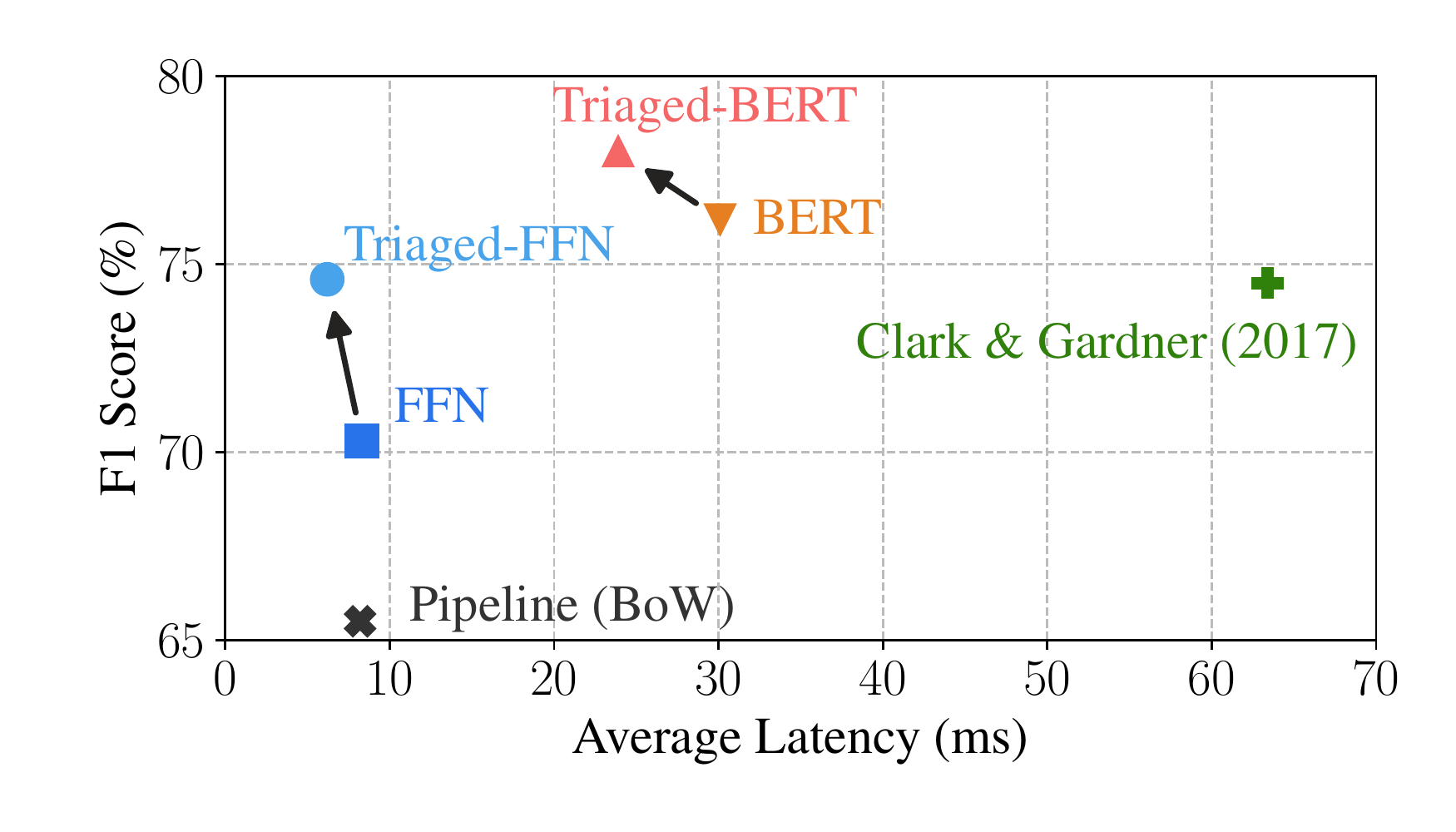}
    \vspace{-2ex}
    \caption{Using a \moduleA{} improves both the speed and accuracy on doc-SQuAD.
    \label{fig:context_pruning_effect}
    }
    \vspace{-0.1in}
\end{wrapfigure}

\subsection{ Main Results with Doc-SQuAD} 
We evaluate  the effectiveness of \method{} on the Doc-SQuAD task.
We pick all hyper-parameters and number of paragraphs (selected by the TF-IDF ranker) through cross-validation on the \textit{dev-val} data set. 

\autoref{fig:context_pruning_effect} shows that effect of \method{} on the FFN and BERT models for the doc-SQuAD baseline in terms of F1 retrieval score and latency. Remarkably, \method{} improves both algorithms across \emph{both} metrics. 
\methodffn{} achieves similar F1 score as \citep{Clark2017SimpleAE} while enjoying a $10\times$ speedup.
It is also slightly faster than the Pipeline method with BoW selector.

Adding a \moduleA{} for context pruning also benefits the BERT model, and results in a 1.6\% F1 jump and a 20\% reduction on the average latency. 
If we optimize the number of paragraphs given to our \methodffn{} as well, it can match the F1 score of BERT on \textit{dev-test} by reading more paragraphs (see \autoref{tbl:main_results}). 
Inevitably, reading longer contexts slows down the model. 

The pipeline methods, though designed to be very efficient don't perform well on doc-SQUAD. We believe the reason is that doc-SQuAD could be too challenging. Compared to the three datasets used in \citep{choi2017coarse}, the questions and answers in SQuAD are longer and require more complex reasoning. Moreover, SQuAD has fewer training examples compared to these datasets. The pruned portion of the pipeline methods is relative low because we tune the number of sentences to keep for the best F1 score.

\subsection{Ablation Studies}

\begin{figure}[t]
    \centering
    \begin{minipage}{0.45\textwidth}
        \vspace{-2ex}
        \centering{\includegraphics[width=\textwidth]{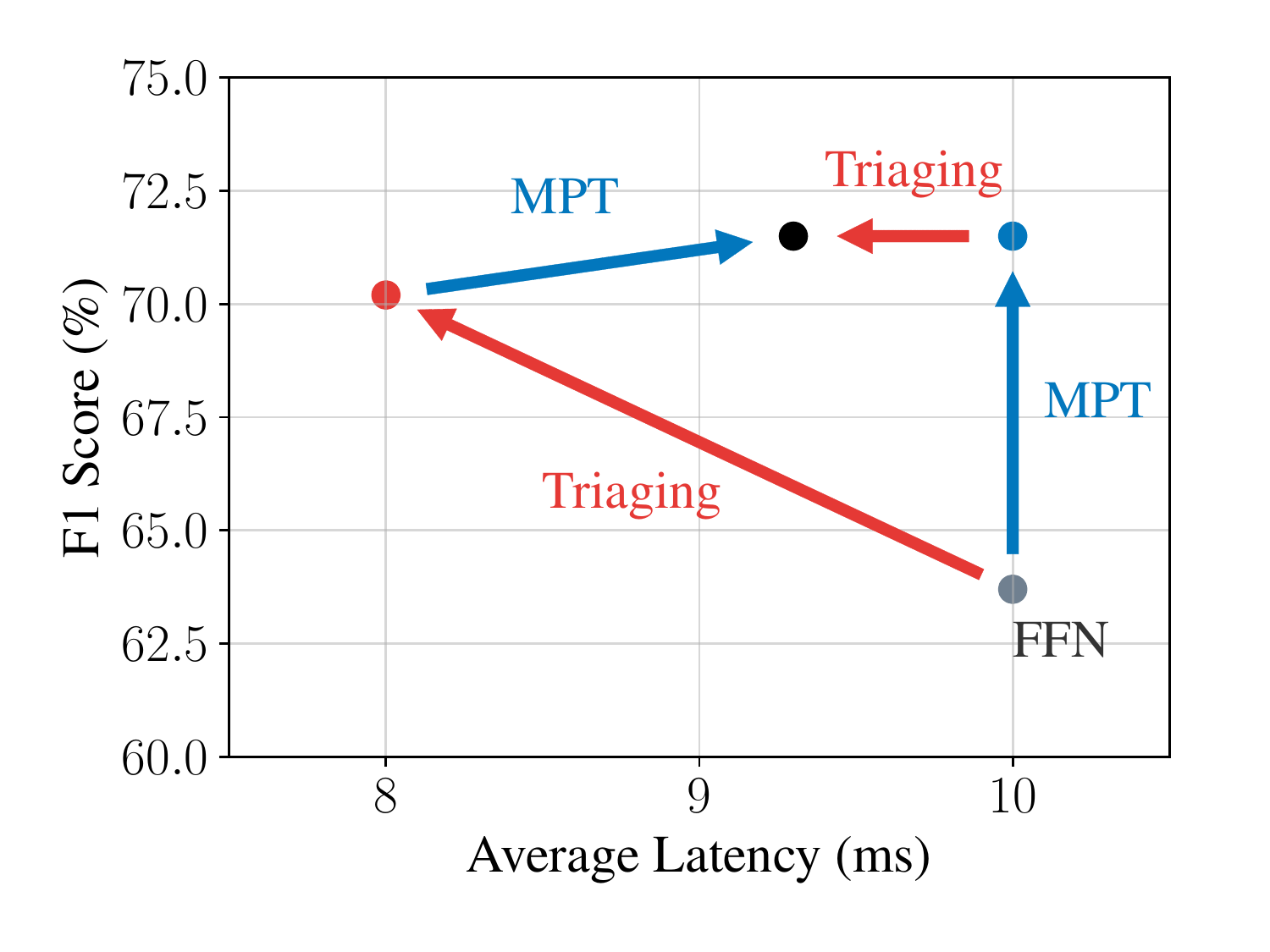}}
        \vspace{-1ex}
        \caption{\method{} reduces the average latency no matter whether multi-paragraph training method     \citep{Clark2017SimpleAE} is used or not.
        }
        \label{fig:ablation}
    \end{minipage}\hfill
    \begin{minipage}{0.45\textwidth}
        \vspace{-2ex}
        \centering{\includegraphics[width=\linewidth]{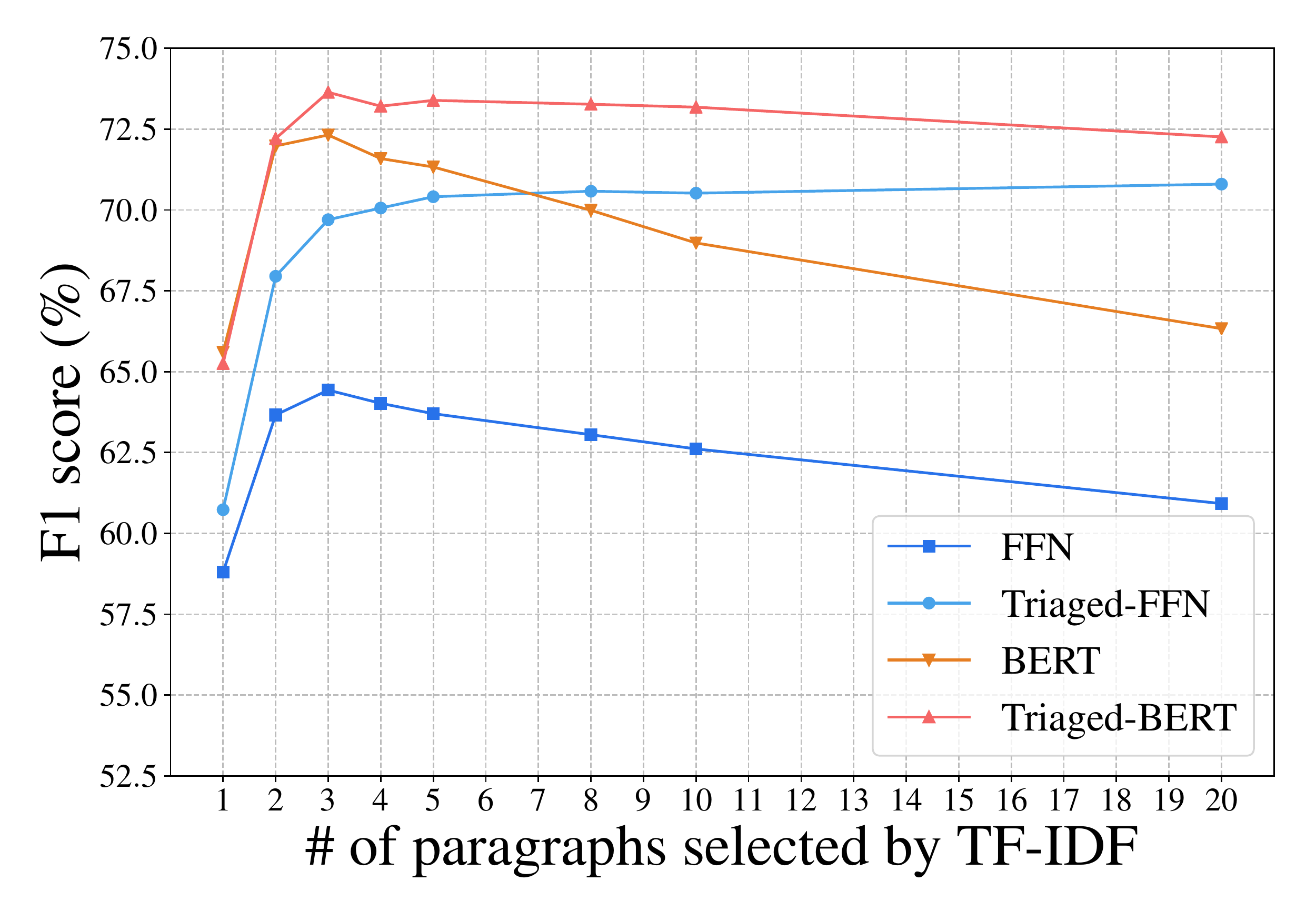}}
        \vspace{-.2in}
        \caption{
        Triaged models
        ({\color{triagedbert}red} and {\color{triagedffn}cyan}) are more robust to long input contexts. In contrast, the performance of models without triaging ({\color{bert}orange} and {\color{ffn}blue}) begins to deteriorate as the context length increases.
        }
        \label{fig:doc_squad_varying_n_para}
    \end{minipage}
    \vspace{-.1in}
\end{figure}

\begin{table*}[]
\centering \small
\begin{tabular}{l|rr|rrr|r}
\multirow{2}{*}{Method} & 
\multicolumn{2}{c|}{\textit{dev-val}} &
\multicolumn{3}{c}{\textit{dev-val\ } Latency (ms)} &
\multicolumn{1}{|l}{Pruned}
\\
& 
\multicolumn{1}{l}{EM} &
\multicolumn{1}{c|}{F1} &
\multicolumn{1}{c}{Avg.} &
\multicolumn{1}{c}{90\%} &
\multicolumn{1}{c}{99\%} &
\multicolumn{1}{|l}{Portion}
\\
\hline
\hline
BERT & 65.4 & 71.3 & 51.6$\pm$0.13 & 74.1$\pm$0.31 & 160.7$\pm$0.45 & 0\%\\
\methodbert{} ($t = \infty$, WS) & 65.5 & 71.8 & 31.5$\pm$0.17 & 39.1$\pm$0.28 & 68.2$\pm$0.60 & 84.2\%\\
\methodbert{} ($t = \infty$) & \textbf{67.4} & \textbf{73.4} & 31.3$\pm$0.32 & 38.7$\pm$0.34 & 70.5$\pm$0.27 & 85.1\%\\
\methodbert{} ($t = 0.6$) & 65.0 & 71.3 & \textbf{26.9$\pm$0.13} & \textbf{36.4$\pm$0.15} & \textbf{63.2$\pm$0.61} & 90.1\% \\

\hline
FFN & 56.6 & 63.7 & 10.0$\pm$0.18 & 12.1$\pm$0.22 & 19.4$\pm$0.59 & 0\% \\
\methodffn{} ($t=\infty$) & \textbf{62.3} & \textbf{70.4} & 10.1$\pm$0.03 & 11.8$\pm$0.10 & 21.9$\pm$0.09 & 85.6\% \\
\methodffn{} ($t=0.4$) & 62.2 & 70.2 & 8.0$\pm$0.11 & 11.3$\pm$0.27 & 21.2$\pm$0.20 & 90.8\% \\
\methodffn{} ($t=0.1$) & 60.1 & 68.2 & \textbf{6.0$\pm$0.21} & \textbf{10.5$\pm$0.43} & \textbf{17.3$\pm$1.45} & 95.7\% \\

\end{tabular}
\caption{Ablation study on the \textit{dev-val} set of doc-SQuAD with top 5 paragraphs selected by TFIDF cosine similarity. 
$t$: early-exit threshold, $t=\infty$: no early-exit, WS: weight-sharing between the Triage module and the output layer. We fix the number of candidates to 20.
For latency, we report the mean and standard deviation over 5 runs. (Best results for each MRC model are in \textbf{bold}.)}
\label{tbl:full_ablation_trim}
\vspace{-0.1in}
\end{table*}

In this section we investigate various hyper-parameter trade-offs and design decisions for \method{}. 
To balance speed and precision we fix the number of paragraphs to 5 throughout this section.
\autoref{fig:ablation} shows the trade-off between using our method and the multi-paragraph training  (MPT) method introduced in \citep{Clark2017SimpleAE}. Both methods can improve the document-level performance significantly. Though MPT performs slightly better, it doesn't reduce the inference latency. Additionally, it requires the model to see twice as much data during training. Further, the two methods are complementary and can both be applied jointly to maintain high accuracy and efficiency.

The first trade-off we investigate is how many paragraphs one should select with the TF-IDF pre-screening. More paragraphs slow down the inference process but lower the risk of accidentally discarding the correct paragraph in the initial pre-screening phase.   
\autoref{fig:doc_squad_varying_n_para} shows that \methodffn{} and \methodbert{} are more robust to long context compared to their counterparts. The performance of \methodffn{} continue increasing up to 20 paragraphs. Admittedly, including a TF-IDF paragraph ranker as an initial pre-screening step is quite helpful for the overall performance.

We also investigate if it is advantageous if the \moduleA{} shares weights with the output layer of the MRC. 
\autoref{tbl:full_ablation_trim} shows the results with and without weight sharing (Triaged-BERT with and without \emph{WS}). 
The performance with weight sharing drops a little bit compared to its counterpart without weight-sharing (71.8\% vs. 73.4\% F1). However, compared to the baseline MRC model, BERT (71.3\% F1), it does not introduce any additional parameters and obtains a $0.5\%$ F1 boost and a $63\%$ speedup for free.
Finally, \autoref{tbl:full_ablation_trim} also shows the effect of varying early-exit thresholds $t$. 
Without early-exits, $t=\infty$ the accuracy is the highest, but it is also (predictably) the slowest. Significant speed-ups can be obtained through lower values of $t$ (e.g. a reduction from 10ms down to 6ms in the case of FFN). 
\section{Conclusion}
In this paper we introduce \method{}, a novel approach for Machine Reading Comprehension, which incorporates a \moduleA{} between two early layers of a deep MRC model.
The \moduleA{} skims the document and reduces it to a small subset of candidate sentences, and the following layers can focus on these parts and extract high precision answers out of these candidates. 
By pruning the context, the \moduleA{} corrects the covariate shift caused by the mismatch of context lengths between training and testing phases --- 
essentially mimicking the training setting at testing time for the subsequent layers.
We show that this approach has a similar regularizing effect as training on multiple paragraphs, but is much faster during training and testing. 
Finally, the two-stage nature of \method{} allows us to reduce the average latency time further through early-exiting of easy question, context pairs. 

\section*{Acknowledgments}
This research is supported in part by grants from the National
Science Foundation (III-1618134, III-1526012, IIS1149882,
IIS-1724282, and TRIPODS-1740822), the Office
of Naval Research DOD (N00014-17-1-2175), and the
Bill and Melinda Gates Foundation. We are thankful for
generous support by SAP America Inc.

{\small
\bibliography{ref}
\bibliographystyle{plainnat}
}

\appendix
\section{Implementation Details}
\subsection{Experimental setup}
All the experiments are conducted on a single NVIDIA GTX-1080Ti GPU using single precision floating-point. We implement our models under the PyTorch framework \footnote{https://pytorch.org/}.  We do not include the standard data pre-processing time and the post-processing of mapping the position predictions back to answer texts in any experiments to focus on the time spent on the models.

\paragraph{Hyper-parameter tuning.}
For all models, we tune the number of paragraphs in $\{1, 2, 3, 4, 5, 8, 10, 20, 30, 50\}$, the number of candidates $K$ in $\{1, 2, 3, 4, 5, 10, 20, 30, 50\}$, and the early-exit threshold $t$ in $\{0.4, 0.5, 0.6, 0.7, 0.8, 0.9\}$ based on the F1 score on \textit{dev-val}.

\subsection{Baseline details}

\paragraph{Pipeline models.}
\citet{choi2017coarse} proposed three methods to combine a lightweight sentence selector model with a sentence-level sequence-to-sequence question answering model on document-level question answering tasks.
However, we use an FFN trained on sentence-level SQuAD as the answer generator instead because \citep{wang2017matchlstm} have demonstrated that the span prediction models outperform the sequence-to-sequence models by a large margin on SQuAD.
Among three methods proposed by \citep{choi2017coarse}, we focus on the pipeline method in which the sentence selector is trained separately. In their paper, this method usually produces the best sentence selection accuracy, which fits the span prediction the best.
We consider the following 5 choices of the sentence selectors as the baselines.
\begin{itemize}
    \item \textbf{BoW Model:} Proposed by \citet{choi2017coarse}, it concatenates the average of the word embedding of the context and the question to a multi-layer perceptron (MLP) with one hidden layer and softmax-normalize the logits across the sentences to model the probability of a sentence being the golden sentence. 
    \item \textbf{CNN:} Proposed by \citet{choi2017coarse}, instead of taking the average of the word embedding, a convolution with kernel size 5 is performed followed by a temporal max-pooling to summarize the question and the sentence. Again, the summary vector is then fed to a MLP and a softmax layer to produce the result.
    \item \textbf{2-layer Bidirectional SRU:} To have a fair comparison, we equip the sentence selector model with the same building block as our \moduleA{}. A temporal max-pooling, a MLP, and a softmax layer are also used.
    \item \textbf{First $n$ sentences:} We use the TF-IDF ranker to sort the paragraphs and choose the first $n$ sentences in this reordered document.
    \item \textbf{Random Selector:} We report the performance of randomly chosen sentences as yet another baseline.
\end{itemize}
We also use the 300-d GloVe pre-trained embedding\footnote{\url{http://nlp.stanford.edu/data/glove.840B.300d.zip}} to initialize these sentence selector models. 
Because the documents can be too long to fit into a GPU, in each epoch, we sample a paragraph and concatenate it with the golden paragraph to train the model. 
We try two settings: i) sampling the paragraph from the whole document and ii) sampling it from the top 4 relevant paragraphs selected by the TF-IDF ranker to the question. Based on the F1 score on \textit{dev-val}, sampling paragraphs from the whole document produces slightly better results.

\paragraph{BiDAF + Self-attn + MPT}~\citep{Clark2017SimpleAE} is the state-of-the-art model on doc-SQuAD which is an improved version of BiDAF~\citep{seo2016bidirectional} that uses an additional self-attention module to improve the accuracy and replaces LSTMs with GRUs~\citep{cho2014gru} for faster speed.
It employs their multi-paragraph training (MPT) method which samples two paragraphs among the top 4 relevant paragraphs (based on TF-IDF cosine similarity) and the golden paragraph in each epoch. We use the open-sourced code\footnote{\url{https://github.com/allenai/document-qa}} provided by the authors.

\paragraph{BERT question answering model} is the state-of-the-art model on the standard SQuAD dataset, which demonstrates the limit of applying a paragraph-level SQuAD model to doc-SQuAD. We use the PyTorch open source implementation\footnote{\url{https://https://github.com/huggingface/pytorch-pretrained-BERT}} with default hyperparameters.

\paragraph{FastFusionNet (FFN)}~\citep{wu2019fastfusionnet} is an efficient machine reading comprehension model. We use the open source\footnote{\url{https://github.com/felixgwu/FastFusionNet}} with default hyperparameters.

\subsection{Triaged MRC Models}
For \methodbert{}, we use the same set of hyperparameters as well as the same pre-processing as BERT.
Similarly, we use the same set of hyperparameters as well as the same pre-processing as FFN for \methodffn{}.

\begin{table*}
\vspace{-2ex}
\centering \small
\begin{tabular}{l|rll}
\hline
 & \begin{tabular}[c]{@{}l@{}}1-example Latency
\end{tabular} & Model     & Hardware    \\
\hline
\multirow{4}{*}{Triaged model}  
& \textbf{2.9 ms} & \moduleA{} of \methodffn{} (F1 78.3) & 1 GTX-1080 Ti\\
  & 3.9 ms     & \methodffn{} + EE($t$=0.2) (F1 80.9) & 1 GTX-1080 Ti \\
  & 6.2 ms     & \methodffn{} + EE($t$=0.6) (F1 82.6)  & 1 GTX-1080 Ti \\
  & 7.5 ms     & \moduleA{} of \methodbert{} (F1 76.1)  & 1 GTX-1080 Ti \\
\hline
 \multirow{4}{*}{\begin{tabular}[c]{@{}l@{}}Previous\\ Work\end{tabular}}  
  & 7.6 ms     & PA-Occam-Bert (F1 75.9) & 1 V100 \\
  & 7.9 ms     & FFN (F1 82.5)  & 1 GTX-1080 Ti \\
  & 22.3 ms     & BERT (F1 88.5)  & 1 GTX-1080 Ti \\
  & 45.5 ms     & FusionNet (F1 83.6)  & 1 GTX-1080 Ti \\
  & 100.0 ms    & BiDAF (F1 77.3)   & 16 CPU    \\
  & 590.0 ms    & BiDAF (F1 77.3)    & 1 K80    \\
  & 638.1 ms      & BiDAF (F1 77.3)    & 1 P100     \\
\hline
\end{tabular}
\caption{DAWNBench Inference time of
\method{}
}
\label{tbl:dawnbench}
\vspace{-3ex}
\end{table*}

\section{More Experiments}

\subsection{TriviaQA with Distant Supervision}
We further evaluate our context pruning method on the TriviaQA~\citep{joshi2017trivia} where the golden paragraphs are not available; instead, the documents containing the ground truth answers are selected based on distant supervision. We train a smaller version of FFN and \methodffn{} on TriviaQA web split and tune the number of candidates of the context pruning on the verified development set. With context pruning, the \methodffn{} gets 1.8$\times$ and 3.0$\times$ speedup on average latency and 99 percentile latency respectively compared to FFN, while only sacrificing 0.7 \% F1 score on the test set (from 69.1 \% to 68.4 \%). 

\subsection{DAWNBench} 
DAWNBench is a benchmark suite for end-to-end deep learning training and inference~\citep{coleman2017dawnbench}.
We demonstrate the efficiency of our fast models based on two metrics defined by DAWNBench in \autoref{tbl:dawnbench}.
\paragraph{The inference time track} measures the average 1-example inference latency of a model with an F1 score of at least $75\%$. The latency of our \moduleA{} with F1 score $78.3\%$ is only $2.9$ ms (a $34\times$ speedup compared to the $100$ ms BiDAF on the leader-board). A stronger baseline would be to fine-tune the embedding and the first 4 blocks of a pre-trained BERT since the models with Transformer blocks usually have lower latency than the RNN-based models with similar performance. Our model is still 2.6 times as fast as this baseline.
With early-exit, our \model{} gets $80.92\%$ F1 score with $3.9$ ms latency.

\subsection{Easier and harder questions}
\autoref{tbl:confident_trigram} shows the common trigrams to start confident or hesitant questions of the \moduleA{} of a \methodffn{}.
To be specific, we sort the questions by confidence and choose the top 10\% confident ones and count the first three words in these questions. We observe that questions asking for a time especially for a year are often easier.
In contrast, questions starting with ``what" or ``how" are more likely to be hard. 

\begin{table}[h]
\centering

\resizebox{0.5\linewidth}{!}{%
\begin{tabular}{lr|lr}
\multicolumn{2}{l|}{Confident}                      & \multicolumn{2}{l}{Hesitant}                \\ \hline
Trigram                & \multicolumn{1}{l|}{Count} & Trigram         & \multicolumn{1}{l}{Coune} \\ \hline
what be the            & 171                        & what be the     & 190                       \\
in what year           & 98                         & what do the     & 55                        \\
who be the             & 70                         & what type of    & 51                        \\
when be the            & 69                         & what do luther  & 31                        \\
when do the            & 40                         & how do the      & 24                        \\
in which year          & 29                         & what be a       & 21                        \\
what year do           & 27                         & what do tesla   & 19                        \\
what year be           & 16                         & what be one     & 18                        \\
what be another        & 15                         & where be the    & 17                        \\
where be the           & 14                         & where do the    & 14                        \\
how much do            & 11                         & why be the      & 13                        \\
who design the         & 11                         & who be the      & 11                        \\
approximately how many & 9                          & what happen to  & 11                        \\
on what date           & 9                          & when be the     & 11                        \\
when do tesla          & 8                          & what be tesla   & 10                        \\
what percentage of     & 8                          & how do luther   & 10                        \\
what do the            & 8                          & what kind of    & 10                        \\
how old be             & 8                          & what be another & 10                        \\
when do luther         & 7                          & what have the   & 7                         \\
who lead the           & 7                          & how be the      & 7                        
\end{tabular}
}

\caption{Common trigrams to start confident or hesitant questions of the \moduleA{}.}
\label{tbl:confident_trigram}
\end{table}
\label{sec:supplemental}

\end{document}